\pdfoutput=1

\documentclass[11pt]{article}

\usepackage{acl}

\usepackage{times}
\usepackage{latexsym}
\usepackage{bm} 
\usepackage{graphicx}
\usepackage{amssymb}
\usepackage{amstext}
\usepackage{multirow,multicol}
\usepackage{threeparttable}
\usepackage{makecell}
\usepackage{booktabs}
\usepackage{bbding}
\usepackage{arydshln} 
\usepackage{stfloats}
\usepackage{enumitem}

\usepackage[T1]{fontenc}

\usepackage[utf8]{inputenc}

\usepackage{microtype}

\usepackage{inconsolata}

\title{Label-Synchronous Neural Transducer for \\ E2E Simultaneous Speech Translation}

\author{Keqi Deng, Philip C. Woodland \\
         Department of Engineering, University of Cambridge, Trumpington St., Cambridge, UK. \\ \texttt{\{kd502, pw117\}@cam.ac.uk} \\}

\begin{document}
\maketitle
\begin{abstract}

While the neural transducer is popular for online speech recognition, simultaneous speech translation (SST) requires both streaming and re-ordering capabilities. This paper presents the LS-Transducer-SST, a label-synchronous neural transducer for SST, which naturally possesses these two properties. The LS-Transducer-SST dynamically decides when to emit translation tokens based on an Auto-regressive Integrate-and-Fire (AIF) mechanism. A latency-controllable AIF is also proposed, which can control the quality-latency trade-off either only during decoding, or it can be used in both decoding and training. The LS-Transducer-SST can naturally utilise monolingual text-only data via its prediction network which helps alleviate the key issue of data sparsity for E2E SST. During decoding, a chunk-based incremental joint decoding technique is designed to refine and expand the search space. Experiments on the Fisher-CallHome Spanish (Es-En) and MuST-C En-De data show that the LS-Transducer-SST gives a better quality-latency trade-off than existing popular methods. For example, the LS-Transducer-SST gives a 3.1/2.9 point BLEU increase (Es-En/En-De) relative to CAAT at a similar latency and a 1.4~s reduction in average lagging latency with similar BLEU scores relative to Wait-k.

\end{abstract}

\section{Introduction}

Simultaneous speech translation (SST) generates translations from input speech in a streaming fashion. Conventional cascaded SST performs streaming automatic speech recognition (ASR) followed by text-based simultaneous machine translation \cite{DBLP:journals/mt/FugenWK07}. Recently, end-to-end (E2E) SST has become popular and has advantages, including lower latency \cite{ren2020simulspeech, Ma2020SimulMTTS}.
However, E2E SST is challenging since it requires taking into account word re-ordering between source and target languages during the streaming process \cite{chang22f_interspeech}. Neural transducers, the dominant model for low-latency ASR, find this difficult due to their monotonic nature. Furthermore, E2E training results in severe data sparsity \cite{Bentivogli2021CascadeVD}.

Current SST methods are normally based on the Transformer \cite{Vaswani2017} attention-based encoder-decoder (AED) structure, which isn't naturally able to deal with streaming, as noted by \citet{Xue2022LargeScaleSE}. A popular approach to adapt the AED to SST is the Wait-k policy \cite{Ma2020SimulMTTS}, which is a fixed read-write policy that uses a fixed number of wait duration steps before translation. However, it can be too aggressive or conservative in different cases \citep{DBLP:conf/acl/ZhengLZMLH20}. Alternative methods involve a flexible policy which lets the model decide how much input to read before generating the next translation token \cite{polak23_interspeech}, such as monotonic multi-head attention (MMA) \cite{Ma2020Monotonic} and Continuous Integrate-and-Fire (CIF) \cite{9054250} based SST system. However, these flexible policies normally rely on a latency loss to adjust the quality-latency trade-off at training, unlike the fixed Wait-k policy that can control latency at decoding only.


To better address the challenges of E2E SST, this paper adapts the label-synchronous neural transducer \cite{deng2023labelsynchronous} developed for ASR to SST and denotes the resulting technique the LS-Transducer-SST. In the LS-Transducer-SST, an Auto-regressive Integrate-and-Fire (AIF) mechanism uses accumulated frame-level weights to dynamically determine 
when to emit translation tokens,
based on which a label-level target-side encoder representation is extracted auto-regressively using an attention mechanism. Therefore, the LS-Transducer-SST is naturally equipped with both streaming and re-ordering capabilities. In addition, the prediction network of the LS-Transducer-SST works as a standard language model (LM) as its output is directly combined with the extracted encoder representation at the label level. As a benefit, the E2E SST data sparsity issue can be alleviated because the prediction network can effectively utilise monolingual text-only data, which is normally easy to collect, for tasks such as pre-training or text-based adaptation. While the standard AIF theoretically ensures low-latency output for the LS-Transducer-SST, to better control the quality-latency trade-off, a latency-controllable AIF is proposed, which controls the latency by adjusting the decision threshold of the accumulated frame-level weights. Furthermore, the latency-controllable AIF allows the quality-latency trade-off to be controlled not only during training but also during decoding, enabling the LS-Transducer-SST to combine the advantages of typical fixed and flexible SST policies. 
This paper focuses on low/medium-latency scenarios to keep the low-latency advantage of E2E SST.
During decoding, to improve translation quality, a chunk-based incremental joint decoding is further proposed to refine and expand the search space.
The proposed LS-Transducer-SST was evaluated on Fisher-CallHome Spanish (Es-En) and MuST-C En-De corpora and gave an improved quality-latency trade-off compared to existing popular SST methods.
The main contributions are summarised below:
\vspace{-0.1cm}
\begin{itemize}
\setlength{\itemsep}{2pt}
\setlength{\parsep}{0pt}
\setlength{\parskip}{0pt}
    \item LS-Transducer-SST, naturally equipped with streaming and reordering abilities, is proposed for SST and can help alleviate its data sparsity.
    \item A latency-controllable AIF is proposed to control the latency during decoding effectively.
    \item A chunk-based incremental joint decoding is proposed to expand the search space.
    \item Extensive experiments were conducted. Our code bridges the ESPnet and Fairseq toolkits and will facilitate future research\footnote{The code is available at: \url{https://github.com/D-Keqi/LS-Transducer-SST}}.
\end{itemize}


\section{Related Work}
\vspace{-0.06cm}
\subsection{Label-synchronous Neural Transducer}
The standard neural transducer \cite{Graves2012SequenceTW} is a frame-synchronous model,
which operates on a frame-by-frame basis and uses blank tokens to augment the output sequence.
However, blank token prediction is inconsistent with the LM task, which means the
prediction network doesn't operate as a standard LM and cannot effectively utilise text data \cite{Chen2021FactorizedNT}. The label-synchronous neural transducer \cite{deng2023labelsynchronous} 
introduces
an Auto-regressive Integrate-and-Fire (AIF) mechanism, which extracts label-level encoder representations that are directly combined with the prediction network at the label level. Therefore, the operation becomes label-synchronous and the prediction network is consistent with the LM task. 

To be more specific, AIF learns frame-level weights $(\alpha_1, \cdots, \alpha_T)$ for each encoder output frame $\mathbf{E}=(\bm{e}_1, \cdots, \bm{e}_T)$. To generate the $i$-th output unit, the weight $\alpha_t$ is accumulated from left to right until it exceeds $i$, and then the time step of this located boundary is denoted as $T_i$+1. Hence, AIF estimates a monotonic alignment for streaming ASR. With this located boundary, the label-level representation $\bm{h}_i^{\rm aif}$ is extracted using dot-product attention with $\textbf{E}_{1:T_j}$ as the keys and values:
 \begin{equation}
    \bm{h}_i^{\rm aif} = {\rm softmax}(\bm{q}_i\cdot{ {\textbf{E}_{1:T_i}}^{\top}})\cdot{\textbf{E}_{1:T_i}} \label{label}
\end{equation}
where the query $\bm{q}_j$ can be the prediction network intermediate output. Suppose $\bm{h}_i^{\rm pred}$ is the output of the prediction network which normally has the same structure as a Transformer LM, the joint network combines $\bm{h}_i^{\rm aif}$ and $\bm{h}_i^{\rm pred}$ at the label level and computes the predicted logits $\bm{l}_{i}$ as follows:
\begin{equation}
    \bm{l}_{i} = {\rm FC}(\bm{h}_i^{\rm aif})+{\rm FC}(\bm{h}_i^{\rm pred})
\end{equation}
where FC denotes a fully-connected network that maps the dimension to the vocabulary size. Since the $\bm{l}_{i}$ has the same length as the target sequence, the cross-entropy loss can be used as the training objective. In addition, a connectionist temporal classification (CTC) loss is also computed by the encoder to improve model training.

In ASR decoding, streaming joint decoding is used that also considers an online CTC prefix score, which is modified to be strictly synchronised with the AIF alignment.


\subsection{E2E Simultaneous Speech Translation}
\label{e2e-sst}

For E2E simultaneous speech translation (SST), the fixed Wait-k \cite{Ma2020SimulMTTS} policy is still the most popular approach \cite{chang22f_interspeech}. While adopting this fixed policy, there are several studies that try to improve the ``pre-decision" process, including using CTC \cite{ren2020simulspeech, zeng-etal-2021-realtrans} and the Continuous Integrate-and-Fire (CIF) mechanism \cite{dong-etal-2022-learning}. 

\begin{figure}[t]
    \centering
    \includegraphics[width=76mm]{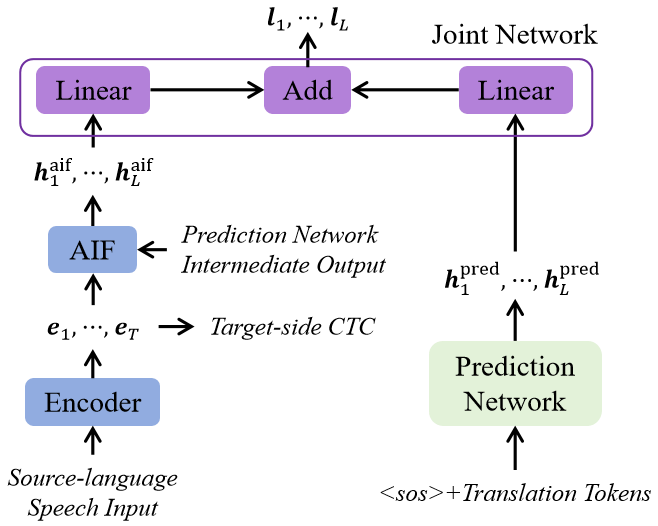}
    \vspace{-0.1cm}
    \caption{Illustration of the proposed LS-Transducer-SST. Linear denotes a linear classifier. Target-side CTC uses translations in the training objective computation.} 
    \vspace{-0.2cm}
    \label{ls-t}
\end{figure}

However, a flexible policy in which decisions to emit translation tokens is made on the fly, is desirable for simultaneous translation \cite{DBLP:conf/acl/ZhengLZMLH20}. A typical flexible policy is monotonic multi-head attention (MMA) \cite{Ma2020Monotonic}, however, \citet{chang22f_interspeech} noted its complex training techniques and proposed a flexible method CIF-IL, in which CIF is used directly to estimate when to output a translation token instead of being used as a pre-decision module. Given that CIF is a monotonic method with limited re-ordering capabilities, a Transformer decoder is built on top of the CIF output, in which infinite lookback (IL) cross-attention attends to the previous CIF output.
However, CIF suffers from mismatched testing and training.

\citet{Xue2022LargeScaleSE} directly explored using a standard neural transducer 
for the SST task. However, the standard neural transducer suffers from limited re-ordering capabilities due to its monotonic alignment \cite{liu-etal-2021-cross}. To address this, the CAAT technique \cite{liu-etal-2021-cross} augments it with cross-attention to remove the strong monotonic constraint, but leads to the exacerbated issue of using large memory during training.
Similarly, \citep{tang-etal-2023-hybrid} combined the Transducer and AED (TAED) by using the AED decoder as the prediction network for the Transducer. However, during training, the number of forward computations for the AED decoder needs to increase in proportion to the length of input speech. 
CAAT and TAED, to keep the streaming property while handling re-ordering, complicate the training due to the frame-synchronous nature, which is an important difference from our label-synchronous approach.

Recently, several papers \cite{liu20s_interspeech, papi-etal-2023-attention, papi23_interspeech} used offline-trained AED models for SST inference. 
Local Agreement (LA) \cite{liu20s_interspeech} generates two consecutive hypotheses and takes agreeing prefixes as the stable hypothesis. \citet{papi-etal-2023-attention} noted that this strategy affects latency and proposed to use encoder-decoder attention (EDA{\small TT}) to decide when to emit translations.

However, here we focus on low and medium-latency scenarios. Since the offline-trained models have a more serious mismatch when decoding with low and medium latency, here we focus on training models in a streaming manner.

\section{Proposed Method: LS-Transducer-SST}
\label{method}
The LS-Transducer-SST, as shown in Fig.~\ref{ls-t}, has four key components: encoder, AIF, prediction network, and joint network. The encoder and AIF extract label-level target-side representations $(\bm{h}_1^{\rm aif}, \cdots, \bm{h}_L^{\rm aif})$ from the source-language speech in a streaming fashion, while AIF controls the time steps at which the representations are emitted.

In addition, the prediction network is an auto-regressive structure, e.g. Transformer LM structure, which generates representations $(\bm{h}_1^{\rm pred}, \cdots, \bm{h}_L^{\rm pred})$ based on previous translation tokens. Since the representations obtained from AIF and the prediction network have the same length, the joint network can directly add the logits obtained from them using linear fully-connected layers, and the prediction network performs as an explicit LM. 

Hence, the output of the LS-Transducer-SST is a 2-dimensional matrix $\mathbb{R}^{L\cdot V}$ ($V$ is the vocabulary size), 
which can use a cross-entropy loss computed with target translations for training
and resolve the issue of expensive training for the standard neural transducer, whose output is a 3-dimensional tensor.

\subsection{Latency-controllable AIF}

\begin{figure}[t]
    \centering
    \includegraphics[width=77mm]{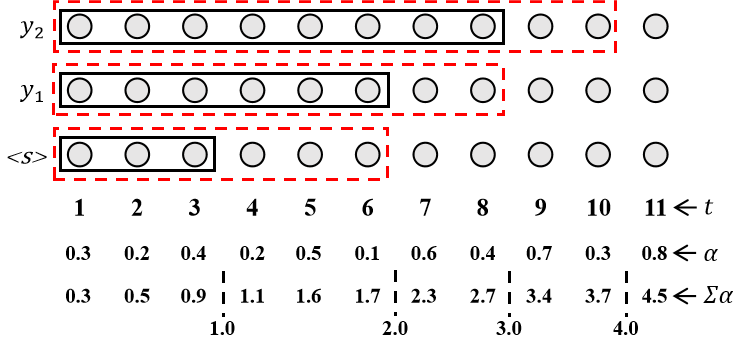}
    \caption{Illustration of latency-controllable AIF. $t$ denotes the time step. $\alpha$ is the frame-level weight. The black solid line shows when the tokens are emitted under standard AIF; the red dotted line illustrates the case when the AIF decision threshold is increased by 1.}
    \label{aif}
\end{figure}

\begin{figure*}[t]
    \centering
    \includegraphics[width=152mm]{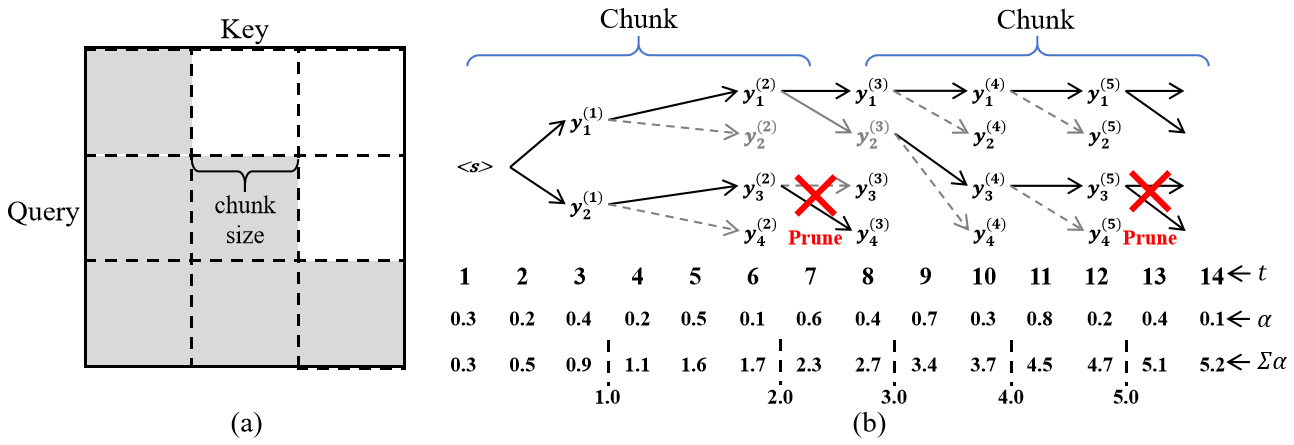}
    \vspace{-0.35cm}
    \caption{Illustration of the proposed chunk-based incremental joint decoding. (a) an illustration of the chunk-based mask; (b) an example of the chunk-based incremental pruning according to the accumulated AIF weights $\sum \alpha$, in which the chunk size is 7, the beam size is 2 within a chunk, the decision threshold of the $i$-th output $y^{(i)}$ is $i$.}
    \vspace{-0.1cm}
    \label{decode}
\end{figure*}

In the LS-Transducer-SST, AIF computes frame-level weights $(\alpha_1, \cdots, \alpha_T)$ for each encoder output frame $\mathbf{E}=(\bm{e}_1, \cdots, \bm{e}_T)$ to dynamically decide how much input to read before emitting the next translation token giving a flexible policy. Following \cite{deng2023labelsynchronous}, the last element $e_{t,d}$ of each frame $\bm{e}_t$ was used as the raw scalar attention value to compute $\alpha_t$.
To mitigate overly sharp weights, this paper proposes a smoothing process to compute the weights using a smoothing factor $\delta$ as follows:
\begin{equation}
\alpha_t = (1-\delta) \times{\rm sigmoid}({e_{t,d}})+\delta \label{smooth}
\end{equation}
where $d$ is the dimension size of $\bm{e}_t$. 
To decide when to emit the $i$-th translation token, 
$\alpha_t$ is accumulated from left to right until it exceeds the decision threshold $i$, as shown in the black solid line of Fig.~\ref{aif}. Suppose this time step is $T_i$+1, $\textbf{E}_{1:T_i}$ is used to extract $\bm{h}_i^{\rm aif}$. Note if the decision threshold has not been reached until all the speech has been read, $T_i = T$, i.e. the entire $\mathbf{E}$ is used to extract the $\bm{h}_i^{\rm aif}$. To help AIF learn this cross-lingual speech-text alignment, a target-side CTC branch is computed which can encourage the Transformer encoder to re-order the output according to the target translation sequence \cite{chuang-etal-2021-investigating, deng22b_interspeech}. Furthermore, a quantity loss $\mathcal{L}_{\rm qua}=|\sum_{i=1}^T\alpha_i - L|$ is computed to ensure that the accumulated AIF weight $\sum_{i=1}^T\alpha_i$ approaches the target translation token length $L$.

After obtaining the output time step $T_i$, 
in contrast to
\cite{deng2023labelsynchronous} that uses simple dot-product attention to generate representations for the ASR task, preliminary experiments showed that multi-head attention is more effective for the
SST task, so $\bm{h}_i^{\rm aif}$ is computed as:
 \begin{equation}
    \bm{h}_i^{\rm aif} = \text{Multihead-attention}(\bm{q}_i, \textbf{E}_{1:T_i}, \textbf{E}_{1:T_i}) 
\end{equation}
where $\textbf{E}_{1:T_j}$ is used as the keys and values, and the query $\bm{q}_i$ is the prediction network intermediate output at the $i$-th step as shown in Fig.~\ref{ls-t}. 

The AIF mechanism provides a natural approach to control the latency by adjusting the decision threshold. In the standard AIF, the decision threshold of the $i$-th translation token is $i$. By adding a hyper-parameter $\epsilon$ into the decision threshold, i.e. $i+\epsilon$, the quality-latency trade-off can be controlled, which is called the latency-controllable AIF.
As shown in Fig.~\ref{aif}, the red dotted line represents the case of $\epsilon=1$, in which more input speech will be read before deciding to output the $i$-th translation token, thus the translation quality improves at the cost of increased latency.

The latency-controllable AIF has many advantages over conventional E2E SST systems. First, it only uses one hyper-parameter $\epsilon$ to achieve fine-grained latency control and can meet any latency requirements, because $\epsilon$ is not limited to integer values. Compared to the fixed Wait-k policy \cite{Ma2020SimulMTTS} that normally needs to set two hyper-parameter values, including the pre-decision ratio and $k$, the latency-controllable AIF is easier to tune. Second, in contrast to a typical flexible policy\footnote{We note some flexible policies applied to offline-trained models \cite{papi-etal-2023-attention, papi23_interspeech} can control latency in decoding.} that uses a latency loss to control the quality-latency trade-off during training \cite{Ma2020SimulMTTS, chang22f_interspeech}, the latency-controllable AIF can control the latency at decoding time while only requiring a single trained model.

\subsection{Chunk-based Incremental Joint Decoding}
For the ASR task, the label-synchronous neural transducer \cite{deng2023labelsynchronous} is decoded based on beam search. However, for the SST task,  beam search
re-ranks the top hypotheses while reading the input speech, making it hard to process the translation results and evaluate the latency. 
Hence, this paper proposes chunk-based incremental joint decoding, which prunes the hypotheses to the same prefix within a chunk.
The chunk-based streaming Transformer is reviewed in this section before describing chunk-based incremental pruning.
\subsubsection{Chunk-based Streaming Transformer}
This paper uses the Chunk-based \cite{li20_interspeech} Transformer encoder to achieve streaming, which uses a chunk mask to limit the range of query-key dot products for each frame within the Transformer self-attention. As shown in Fig.\ref{decode}, the chunk mask
allows the query to be computed 
only with the keys from the current and previous (history) chunks.


\subsubsection{Incremental Pruning within Chunk}

Since the emitted translation tokens inside a chunk actually all correspond to the same chunk speech duration, 
in order to expand the search space without adding extra latency, beam search can be used within a chunk while selecting only the highest-scoring hypothesis after outputting the last token of this chunk. However, it is not always feasible to know whether a token is the last one in a chunk, i.e., to know in advance if the speech input required for the next token will exceed the range of this chunk.

In the LS-Transducer-SST, the AIF mechanism uses frame-level weights $\alpha$ to decide whether to output a translation token or not, so comparing the accumulation of frame-level weights up to the current chunk with the decision threshold of the next translation output token, it can be confirmed if a token is the last one in a chunk. As shown in the example in Fig.~\ref{decode}, the accumulated weights $\sum_{t=1}^{t=7} \alpha_t$ up to the first chunk is $2.3$, so in the standard AIF case, when outputting the second token $y^{(2)}_{j}$ ($j$ can be e.g. $1, \cdots, 4$ in Fig.~\ref{decode}), it is known that the decision threshold for the next token (i.e. $3$) cannot be reached within this chunk. Hence, pruning is required when emitting the $2$nd translation token, i.e. only the highest-scoring hypothesis is kept. A similar situation occurs with the $5$th token in Fig.~\ref{decode}.


\subsection{Training}
\label{lst-train}
During training, as mentioned in Sec.~\ref{method}, the LS-Transducer-SST uses the cross-entropy (CE) loss $\mathcal{L}_{\rm CE}$ as the training objective since the joint network output is a 2-dimensional matrix. In addition, a target-side CTC branch and the AIF quantity loss $\mathcal{L}_{\rm qua}$  are also computed. Hence, the final training objective of the LS-Transducer-SST is as follows:
 \begin{equation}
    \mathcal{L}=\beta \mathcal{L}_{\rm CTC}+(1-\beta) \mathcal{L}_{\rm CE}+\gamma \mathcal{L}_{\rm qua}\cdot L \label{obj}
\end{equation}
where $L$ is the target translation token length as $\mathcal{L}_{\rm qua}$ is a sentence-level loss. 
$\beta$ is the target-side CTC weight and $\gamma$ is the weight of the $\mathcal{L}_{\rm qua}$.

Since the LS-Transducer-SST prediction network works as a standard LM, it can be initialised by a pre-trained target-language LM before SST training. If an unseen domain is encountered during decoding, target-language target-domain text can be used to fine-tune the prediction network, giving flexible domain adaptation. Since monolingual text is normally easy to collect, the LS-Transducer-SST can help alleviate data sparsity issues in E2E SST.

\subsection{Summary}
In summary, this paper enhances the re-ordering capability of the label-synchronous neural transducer by introducing a target-side CTC branch, 
enabling AIF to decide when to emit translation tokens in SST tasks. 
A chunk-based incremental joint decoding is proposed to meet SST requirements,
which prunes hypotheses within a chunk while expanding the search space.
To flexibly control the quality-latency trade-off, a latency-controllable AIF is proposed that can be used at the decoding stage. 
With these contributions, the LS-Transducer-SST becomes a natural SST method combining the advantages of typical fixed and flexible SST policies.

\section{Experimental Setup}
\label{setup}
\subsection{Dataset}
SST models were trained on the
Fisher-CallHome Spanish (FCS) \cite{post-etal-2013-improved} and MuST-C v1.0 \cite{di2019must} English-German (En-De) 
datasets. The dev/test sets of Europarl-ST \cite{jairsan2020a} Spanish-English (Es-En) were used as cross-domain test sets for the FCS corpus. The monolingual source-domain text-only data for FSC was the training set English translations and Fisher \cite{cieri-etal-2004-fisher} transcriptions. For the MuST-C En-De, the training set German translations and German text from TED2020 \cite{reimers-2020-multilingual-sentence-bert} were used. 
The English text from Europarl-ST was used as the target-domain text.
More data details are listed in Appendix~\ref{sec:appendix:data}.

SST Experiments were implemented based on the ESPnet-ST \cite{inaguma-etal-2020-espnet} toolkit.
To evaluate the latency, SimulEval \cite{ma-etal-2020-simuleval} was used to measure the speech version of the word-level Average Lagging (AL) \cite{Ma2018STACLST, Ma2020SimulMTTS}. 
Detokenized BLEU \cite{papineni2002bleu} results are reported to evaluate translation quality.


\subsection{SST Model Descriptions}
SST models built in this paper have a streaming wav2vec2.0 \cite{DBLP:conf/nips/BaevskiZMA20} encoder\footnote{Note: Section~\ref{related} has a different setup to aid comparisons.}, which was fine-tuned with a chunk-based mask of 64 chunk size, resulting in 640~ms average latency.
The encoder was first pre-trained for ASR before SST training 
\cite{inaguma-etal-2020-espnet}.
More training and decoding details can be found in Appendix~\ref{hyper}.

\begin{table}[t] \small
  \centering
  \setlength{\tabcolsep}{0.52mm}
  \renewcommand\arraystretch{1.12}
  \begin{tabular}{l | c |c | c c}
    \Xhline{3\arrayrulewidth}
     {SST Models on FCS Corpus}&{test}&evltest&\multicolumn{1}{c|}{AL(s)} &{Latency}\\
    \hline
    ESPnet\cite{inaguma-etal-2020-espnet}&50.9& 19.4&\multicolumn{2}{c}{Offline}\\
    Fast-MD\cite{9687894}& 54.4& 21.3&\multicolumn{2}{c}{Offline}\\
    B-AED \cite{deng22b_interspeech}&47.7&15.3&3.434&High\\
    \hline
    Wait-$5$ w/ 360~ms pre-decision &48.9&19.9&2.129&High\\
    Wait-$4$ w/ 360~ms pre-decision &48.1&20.2&2.073&High\\
    Wait-$3$ w/ 360~ms pre-decision&46.8&19.1&1.710&Medium\\
    \cdashline{1-1}
    Wait-$3$ w/ 280~ms pre-decision&42.0&17.0&1.388&Medium\\
    Wait-$1$ w/ 280~ms pre-decision&35.2&15.4&1.254&Medium\\
    \cdashline{1-1}
    Wait-$3$ w/ 200~ms pre-decision&28.7&13.1&1.166&Medium\\
    Wait-$1$ w/ 200~ms pre-decision&25.2&12.5&0.987&Low\\
    \hline
    CIF-IL with $\lambda_{lat}=0.0$&33.1&13.6&1.103&Medium\\
    CIF-IL with $\lambda_{lat}=0.5$&30.3&12.6&0.942&Low\\
    \hline
    CAAT& 44.7&17.7&0.965&Low\\
    Standard Neural Transducer &37.9&12.0&1.443&Medium\\
    \hline
    Proposed LS-Transducer-SST&46.3&20.1&0.759&Low\\
    \quad with $\epsilon=1$ in AIF &\textbf{47.8}&\textbf{20.8}&\textbf{0.912}&Low\\
    \quad with $\epsilon=2$ in AIF&49.7&20.9&1.089&Medium\\
    \quad with $\epsilon=5$ in AIF&\textbf{51.4}&\textbf{21.2}&\textbf{1.578}&Medium\\
    \Xhline{3\arrayrulewidth}
  \end{tabular}
  \caption{BLEU ($\uparrow$) results on the Fisher-CallHome Spanish (Es-En). Case-insensitive BLEU was reported on Fisher-test (4 references), and CallHome-evltest (single reference). AL ($\downarrow$) was tested on the CallHome-evltest. The latency is divided into low, medium, and high regions with thresholds 1, 2, and 4s \cite{ansari-etal-2020-findings}. This paper focuses on low and medium-latency scenarios.}
  \label{fisher}
\end{table}

The proposed \textbf{LS-Transducer-SST} was built with a 6-layer unidirectional Transformer-encoder-based prediction network. 
The prediction network was initialised by a pre-trained source-domain LM and then fixed during SST training. 
Four different baseline models were implemented.
First, the widely-used \textbf{Wait-k} fixed policy model was built, which followed the Transformer AED structure that contained a 6-layer Transformer decoder. The quality-latency trade-off of the Wait-k was adjusted by varying $k$ and the fixed pre-decision step size.
Second, the flexible policy method \textbf{CIF-IL} (see Sec.~\ref{e2e-sst}) which can performs well in low-latency scenarios \cite{chang22f_interspeech} was built with a 6-layer Transformer decoder. The latency loss from \cite{chang22f_interspeech} was used to control latency during training with a weight denoted $\lambda_{lat}$.
Third, the \textbf{CAAT} \cite{liu-etal-2021-cross} was built, 
the prediction network had the same structure as that of LS-Transducer-SST, and the joint network consisted of 6-layer Transformer decoder with self-attention modules removed. 
Note that knowledge distillation used in \cite{liu-etal-2021-cross} was not employed in the Main Experiments in this paper in order to compare fairly with other specifically built models. 
The decision step size $d$ was set to 32.\footnote{Appendix~\ref{hyper} explains the reasons to set $d$ to a fixed value.}
Finally, a \textbf{standard neural transducer} was built with the same prediction network as the CAAT.

\subsection{LM and Text Adaptation}
The source-domain Transformer LM was trained on the monolingual source-domain text-only data for 50 epochs and further fine-tuned on the cross-domain text for 15 epochs as a target-domain LM. 
If a density ratio \cite{9003790} was used for domain adaptation, the weights for the source-domain and target-domain LMs were 0.3.
When conducting text-only domain adaptation for the LS-Transducer-SST, the prediction network was fine-tuned on the cross-domain text for 25 epochs. 

\section{Experimental Results}
The LS-Transducer-SST was compared to fixed and flexible policy models. 
To maintain the low-latency advantage of E2E SST, we focus on the low and medium latency scenarios, i.e. $\text{AL} < 1~\text{s}$ and $1~\text{s} < \text{AL} < 2~\text{s}$ following \cite{ansari-etal-2020-findings}.

\subsection{Main Experiments}

Table~\ref{fisher} gives the SST results on Fisher-CallHome Spanish (FSC) data, on which our models yield good results compared to recent work. For the popular Wait-k model, a large pre-decision step (e.g., 280~ms or 360~ms in Table~\ref{fisher}) is important to achieve promising translation quality, which makes it suitable for medium or high-latency scenarios while performing poorly when low-latency. In contrast, the flexible policy CIF-IL outperformed the Wait-k model in low-latency scenarios. 
In addition, the standard neural transducer didn't work well on the SST task due its monotonic constraint and limited re-ordering ability. It tends to accumulate more information before emitting translations, resulting in medium latency. However, CAAT, which augments the standard neural transducer with cross attention to remove the monotonic constraint, yielding good BLEU results in low-latency scenarios, outperforming the Wait-k and CIF-IL.


While these existing methods showed good results, the proposed LS-Transducer-SST gave a significantly better quality-latency trade-off with the latency-controllable AIF\footnote{Note unless specifically stated, the latency-controllable AIF refers to being used during both training and decoding.}.
Compared to CAAT, for similar latency (i.e. AL $\approx$ 0.9s), the LS-Transducer-SST obtained a 3.1 BLEU gain. Compared to the best BLEU results for Wait-k in Table~\ref{fisher}, the LS-Transducer-SST gave 
competitive BLEU results at 0.759~s AL latency (1.37~s AL reduction).

\begin{table}[t!] \small
  \centering
  \setlength{\tabcolsep}{0.45mm}
  \renewcommand\arraystretch{1.19}
  \begin{tabular}{l | c c |c | c }
    \Xhline{3\arrayrulewidth}
     SST Models&{COMMON}&{HE}&AL(s)&{Latency}\\
    \hline
    Wait-k \cite{yan-etal-2023-espnet} &18.6&\textbf{--}&6.8&\textbf{---}\\
    TBCA \cite{yan-etal-2023-espnet} &23.5&\textbf{--}&2.3&High\\
    MoSST\cite{dong-etal-2022-learning} &20.0&\textbf{--}&2.742&High\\
    \hline
    Wait-$5$ with 360~ms &22.5&21.9&2.629&High\\
    Wait-$3$ with 360~ms &22.3&21.1&2.127&High\\
    \cdashline{1-1}
    Wait-$3$ with 280~ms &21.0&18.9&1.638&Medium\\
    Wait-$1$ with 280~ms &20.6&17.8&1.280&Medium\\
    \cdashline{1-1}
    Wait-$3$ with 200~ms &17.0&12.8&1.014&Medium\\
    Wait-$2$ with 200~ms &16.5&11.8&0.881&Low\\
    Wait-$1$ with 200~ms &16.3&11.2&0.757&Low\\
    \hline
    CIF-IL with $\lambda_{lat}=0.0$& 19.3&18.2&1.354&Medium\\
    CIF-IL with $\lambda_{lat}=0.3$& 18.5&17.2&0.962&Low\\
    \hline
    CAAT &20.4&18.9&1.078&Medium\\
    \hline
    LS-Transducer-SST&20.8&19.3&0.715&Low\\
    \quad with $\epsilon=1$ in AIF&\textbf{21.4}&\textbf{20.0}&\textbf{0.853}&Low\\
    \quad with $\epsilon=3$ in AIF&23.3&21.3&1.188&Medium\\
    \quad with $\epsilon=5$ in AIF&\textbf{23.8}&\textbf{22.3}&\textbf{1.635}&Medium\\
    \Xhline{3\arrayrulewidth}
  \end{tabular}
  \caption{Case-sensitive BLEU ($\uparrow$) results on MuST-C En-De. AL ($\downarrow$) was tested on the tst-COMMON set. This paper focuses on low and medium-latency regions.}
  \label{tab:must}
\end{table}

The results on the MuST-C En-De are shown in Table~\ref{tab:must} and the conclusions were consistent with that of FSC corpus: 1. the Wait-k model worked well in medium or high-latency regimes while the CIF-IL surpassed it for low-latency case; 2. CAAT greatly outperformed the Wait-k and CIF-IL when AL is around 1~s; 3. the proposed LS-Transducer-SST yielded an improved quality-latency trade-off compared with these existing methods. When the latency was around 1~s AL, the LS-Transducer-SST outperformed CAAT with a 2.9 BLEU gain. Compared to the Wait-5 policy with 320~ms pre-decision step size, the LS-Transducer-SST could give a competitive translation quality with 1.188~s AL latency, resulting in a 1.44~s AL reduction.

Quality-latency trade-off curves corresponding to Tables~\ref{fisher} and \ref{tab:must} are shown in Appendix~\ref{vis_main}. Results with additional metrics are given in Appendix~\ref{expand}.
A visual example in Appendix~\ref{app:case:ende} illustrates the re-ordering capability of the LS-Transducer-SST.

\subsection{Cross-domain Experiments}

The Europarl-ST dev/test sets were used to evaluate the cross-domain performance of the SST models trained on FSC data. As shown in Table~\ref{tab:stcross}, the LS-Transducer-SST performed robustly regarding the latency and translation quality in cross-domain scenarios.
From the source domain (i.e. FCS) to the cross-domain, the AL result of LS-Transducer-SST only increased slightly (from 0.759s to 0.915s), whereas the latency of Wait-k more than doubled (from 2.129s to 4.603s), 
since in cross domain, the number of byte pair encoding (BPE) \cite{gage1994} units for target translations tends to be relatively longer than for the source domain as the BPE model is trained on source-domain text\footnote{Note the modelling unit is BPE, whose output time step needs to be mapped to the word level to measure latency.}.
For example, on the evltest set of FSC, each BPE unit corresponds to an average of 207~ms speech, whereas on the test set of Europarl-ST, this value is 148~ms.
This shows the robust advantage of the LS-Transducer-SST as a flexible policy, as the fixed Wait-k policy tends to be affected by the data distribution. 
A case study is shown in Appendix~\ref{app:case:esen}.



\begin{table}[t] \small
  \centering
  \setlength{\tabcolsep}{1.3mm}
  \renewcommand\arraystretch{1.2}
  \begin{tabular}{l | c c| c| c| c}
    \Xhline{3\arrayrulewidth}
     \multirow{2}{*}{SST Models}&\multicolumn{2}{c|}{FCS}&\multicolumn{3}{c}{FCS$\Rightarrow$Europarl}\\
     \cline{4-6}
     &{evltest}&{AL(s)}&{test}&{dev}&{AL(s)} \\
    \hline
    Wait-$5$ with 360~ms&19.9&2.129&9.4&10.6&4.603\\
    {\ }{\ }+Density Ratio &\textbf{--}&\textbf{--}&10.8&12.3&4.565\\
    \hline
    LS-Transducer-SST&20.1&0.759&10.4&11.7&0.915\\
    +Adapt Pred. Net.&\textbf{--}&\textbf{--}&\textbf{12.5}&\textbf{13.8}&\textbf{0.863}\\
    {\ }{\ }++Shallow Fusion &\textbf{--}&\textbf{--}&12.8&14.3&0.931\\
    \Xhline{3\arrayrulewidth}
  \end{tabular}
  \caption{Cross-domain BLEU ($\uparrow$) results on Europarl-ST Es-En dev/test sets for SST models trained on Fisher-CallHome Spanish (FCS). Pred. Net. denotes prediction network.
  AL ($\downarrow$) was tested on the Europarl-ST test set.}
  \label{tab:stcross}
\end{table}

The LS-Transducer-SST exceeded the BLEU score of Wait-k in cross-domain scenarios, showing its robust generalisation. Moreover, since the LS-Transducer-SST prediction network operates as an explicit LM and can directly use target-language target-domain text for fine-tuning, the cross-domain BLEU results of the LS-Transducer were further improved with an adapted prediction network. Even when Wait-k used the density ratio method \cite{9003790} to subtract a source-domain LM score and add a target-domain LM score during decoding for domain adaptation, there was still a 1.7 BLEU gap with LS-Transducer-SST, which is more flexible as it does not rely on an additional external LM. In addition, shallow fusion \cite{chorowski2015attention} can also be used to further improve the LS-Transducer-SST with the target-domain LM (see Table~\ref{tab:stcross}).

\begin{figure}[t]
    \centering
    \includegraphics[width=76mm]{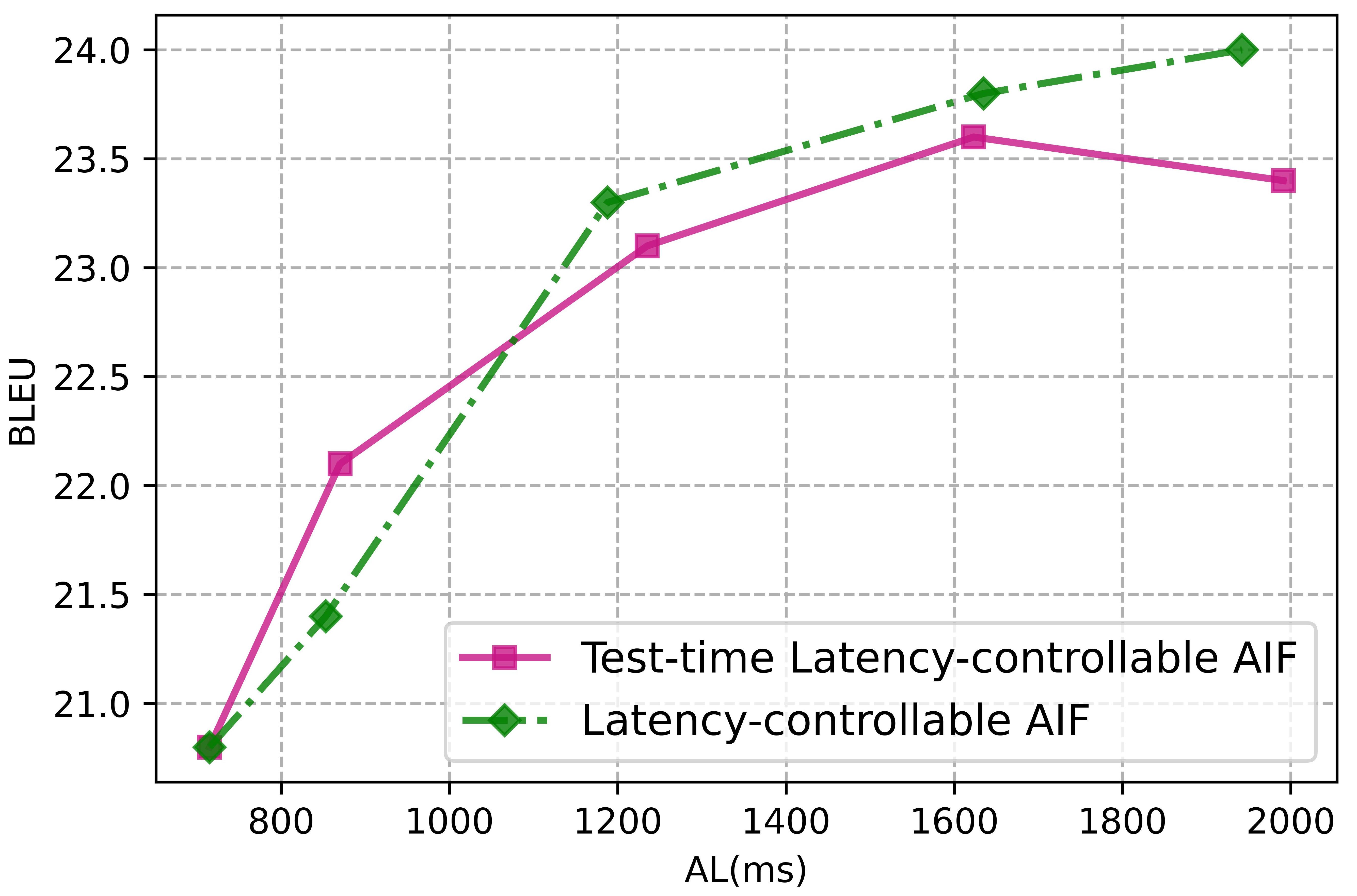}
    \vspace{-0.25cm}
    \caption{Quality-latency trade-off of LS-Transducer-SST on MuST-C En-De tst-COMMON set. The 5 dots for the latency-controllable AIF are $\epsilon \in \{0, 1, 3, 5, 7\}$.}
    \label{tradeoff}
\end{figure}

\subsection{Analysis of Latency-controllable AIF}

Fig.~\ref{tradeoff} shows the quality-latency trade-off curves of the LS-Transducer-SST, in which test-time latency-controllable AIF means using the standard AIF (i.e. $\epsilon=0$) during training and adjusting $\epsilon$ only during decoding. In general, the latency-controllable AIF can flexibly and efficiently control the quality-latency trade-off. While the latency-controllable AIF with consistent training and decoding performed slightly better than the test-time-only one, adjusting $\epsilon$ only in the decoding stage achieves similar results, which is important in real-world deployment as only one model needs to be maintained. Hence, as a flexible policy, the proposed LS-Transducer-SST possesses the advantage of controlling the latency during decoding, which is normally seen in fixed policies like Wait-k.

\begin{table}[t!] \small
  \centering
  \setlength{\tabcolsep}{1.65mm}
  \renewcommand\arraystretch{1.18}
  \begin{tabular}{ l | c c| c }
    \Xhline{3\arrayrulewidth}
     \multirow{2}{*}{SST Models}&\multicolumn{2}{c|}{MuST-C En-De}&\multirow{2}{*}{AL(s)}\\
     &{COMMON}&{HE}& \\
    \hline
    CAAT&20.4&18.9&1.078\\
    {\ } w/ pre-trained pred. net.&18.1&16.5&1.068\\
    LS-Transducer-SST&\textbf{20.8}&\textbf{19.3}&0.715\\
    {\ } w/o pre-trained pred. net.&19.3&18.3&0.704\\
    \Xhline{3\arrayrulewidth}
  \end{tabular}
  \caption{BLEU ($\uparrow$) results for CAAT and LS-Transducer-SST on MuST-C En-De with or without prediction network (abbreviated as pred. net.) pre-training.}
  \label{ablation_pre}
\end{table}

\subsection{Ablation Studies}
The LS-Transducer-SST can effectively utilise monolingual text data by initialising its prediction network with a pre-trained LM. Ablation studies were conducted to evaluate the effectiveness of this initialisation. As shown in Table~\ref{ablation_pre}, the prediction network initialisation was highly effective for the LS-Transducer-SST and was essential to surpass existing methods like CAAT. The monolingual text-only data is normally easier to collect,
which can help alleviate E2E SST data sparsity. However, pre-training the prediction network did not help for CAAT. This is because the CAAT inherits the frame-synchronous property from the standard neural transducer, in which the prediction network is inconsistent with the LM task \cite{Chen2021FactorizedNT}.


Ablation studies on chunk-based decoding and multi-head attention for AIF are in Appendix~\ref{more_ablation}.

\subsection{Comparisons to Recent Work}
\label{related}

To enable further comparisons, the TAED results from \citep{tang-etal-2023-hybrid}, the CAAT results from \citep{liu-etal-2021-cross} and \citep{papi-etal-2023-attention}, and the EDA{\small TT}, LA, and Wait-k results from \citep{papi-etal-2023-attention} were compared to the LS-Transducer-SST in the low/medium-latency scenarios that is our focus.
To allow a fair comparison, in contrast to the Main Experiments, a streaming Conformer \cite{gulati20_interspeech} encoder was employed in the LS-Transducer-SST, and sequence-level knowledge distillation (KD) \cite{kim-rush-2016-sequence} used\footnote{Note only Section~\ref{related} uses the Conformer and KD}. Detailed training settings are given in Appendix~\ref{setting}.

\begin{figure}[ht]
    \centering
    \includegraphics[width=76mm]{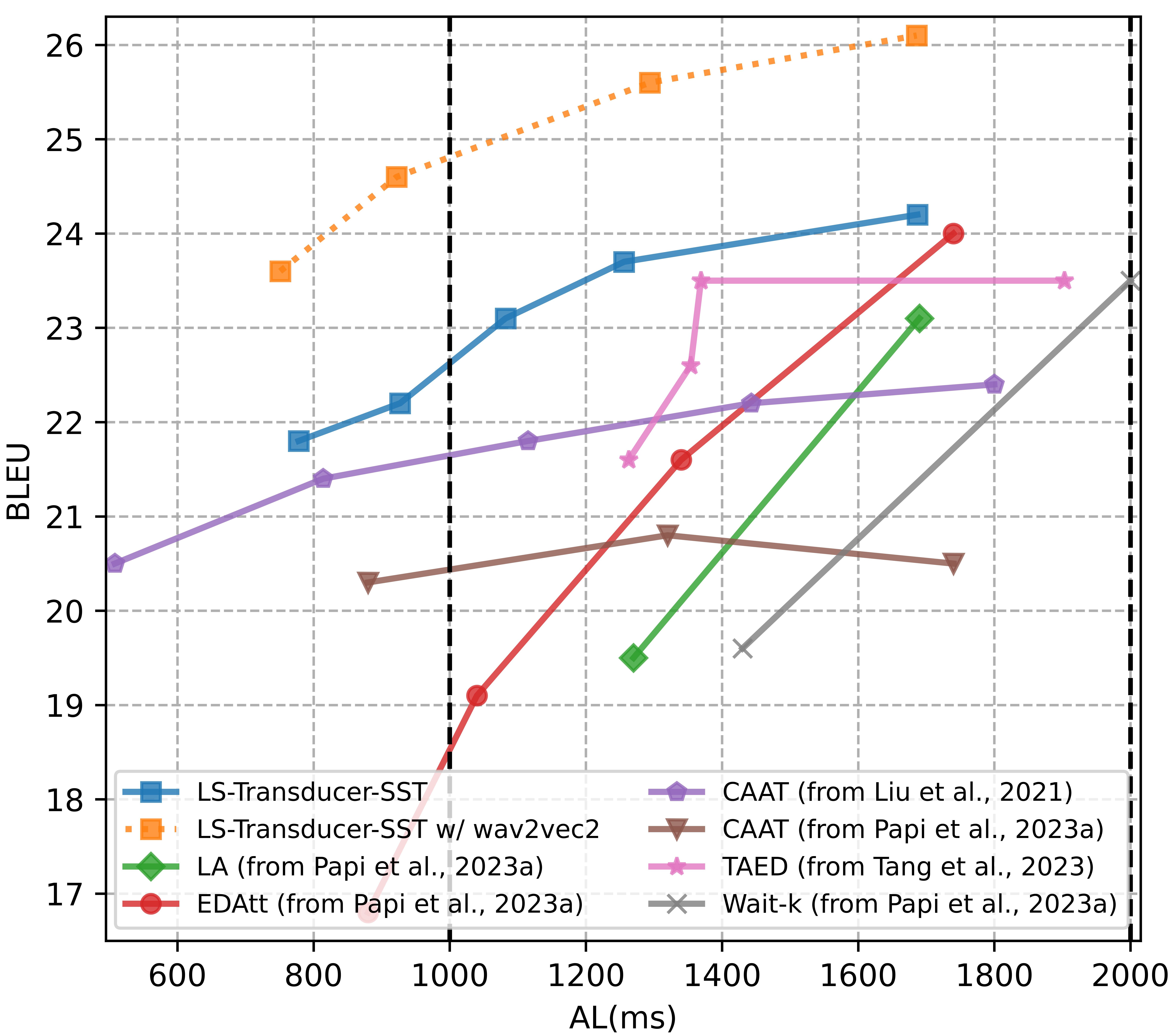}
    \caption{Quality-latency trade-off curves on MuST-C En-De tst-COMMON set. Solid lines are comparable with technique results from the literature. Dotted line indicates the use of wav2vec2.0.
    All results use sequence-level KD in training.}
    \label{related_must}
\end{figure}

As shown in Fig.~\ref{related_must}, CAAT 
shows good results in the low-latency region, while
TAED performs well around 1.4~s. 
In general, CAAT and TAED inherit the neural transducer low-latency advantage, which also motivates our work.
However, the translation quality does not always improve with higher latency for CAAT and TAED.
As latency continues to increase, the mismatch with the offline case reduces, and strategies that use offline-trained models for SST inference, like EDA{\small TT}, begin to surpass other published results. However, LS-Transducer-SST (solid blue line in Fig.~\ref{related_must}) still outperforms other models in both low and medium-latency regions (up to an AL of about 1.7 s). Fig.~\ref{related_must} also shows that a wav2vec2.0 encoder gives further performance benefits.
Online training is especially useful in lower latency operation and  the LS-Transducer-SST is able to effectively adjust the quality-latency trade-off. 
Numerical values  for Fig~\ref{related_must}, along with the additional LAAL metric, are given in Appendix~\ref{numeric}. 

\section{Conclusions}
This paper proposes the LS-Transducer-SST, 
which naturally possesses streaming and re-ordering capabilities. 
By introducing the target-side CTC branch, the re-ordering capability of the label-synchronous neural transducer is enhanced and thus the AIF mechanism can be used for SST. Therefore, the
LS-Transducer-SST is a flexible policy method that dynamically decides when to emit translation tokens using the AIF. 
A latency-controllable AIF is further proposed to effectively control the latency at decoding or training, enabling the LS-Transducer-SST to combine the advantages of typical fixed and flexible SST policies.
In addition, the LS-Transducer-SST provides a natural way to utilise monolingual text-only data, which helps alleviate the E2E SST data sparsity issue. During decoding, a chunk-based incremental joint decoding is further proposed to refine and expand the search space. 
With the focus on low and medium-latency scenarios, experiments showed that the LS-Transducer-SST gives a better quality-latency trade-off than existing methods.

\section*{Limitations}
This paper focuses on low-latency and medium-latency scenarios to maintain the low-latency advantages of E2E SST models,
although some high-latency results are also shown. To explore solutions for high-latency SST, a comparison against cascaded SST systems is worthwhile. However, in real-world deployment, the cascade SST system normally has more available training data than the E2E SST system, such as machine translation and ASR data. The appropriate experimental setup to simulate this real deployment scenario and reasonably compare different SST systems is not straightforward, which is regarded as future work.


Sequence-level knowledge distillation (KD), which uses a neural machine translation model to generate pseudo target text given source text paired with source speech and augments the original data with it \cite{kim-rush-2016-sequence}, was not used in
the Main Experiments (it is used in Section~\ref{related} to allow a close comparison with results from the literature). The reasons are as follows: firstly, this technique requires the source text, i.e. the ASR transcripts, which are not necessarily always available. Secondly, we use the ESPnet-ST toolkit, in which this KD is not the default setting and in general we have followed the standard ESPnet-ST recipes. Thirdly, this method will double the data size, requiring twice the computational resources.
This paper implements four different SST methods to compare with the LS-Transducer-SST, along with adjusting the quality-latency trade-off, making the experiments computationally expensive. Sequence-level KD can be regarded as translation data augmentation and would not be expected to  affect the comparison between different methods. To verify this, we selected some key experiments and give the results after using KD in Appendix~\ref{expand_kd}.

Section~\ref{related} makes further comparisons with recent work from the literature. We have aimed to make it as close a comparison as possible with the literature and hence a Conformer encoder trained on the source speech was used in place of the wav2vec2.0 encoder used in the Main Experiments and also all comparisons use sentence-level KD. However, there are many details which may not be strictly comparable across different papers. For example, the teacher model involved in the sentence-level KD will in general be different between different papers and hence yield different pseudo target text. 
Also, some recent papers have not yet released their code. Furthermore,  even when papers do provide code, others may not always be able to fully reproduce the original results (e.g. the two CAAT results in Fig.~\ref{related_must}). We believe that the comparing to prior published work while attempting to control the precise experimental conditions in Section~\ref{related} and Appendix~\ref{numeric} is the best solution for comparing to the recent papers cited.

The LS-Transducer-SST has so far been evaluated on two European language pairs in this paper, each in a single translation direction (i.e. Es-En and En-De), and while we believe that the technique can be applied to other languages (including non-European languages) and translation directions, the performance has not been verified and is left as future work.

\section*{Ethics Statement}
E2E SST systems suffer from data sparsity issues because speech-translation parallel data is expensive to collect. For under-resourced languages or domains, this issue will be more severe. This can result in a poor user experience for minorities, making minority views to be
under-represented or misunderstood. The LS-Transducer-SST proposed in this paper could be beneficial to this concern as it provides a natural approach to utilise monolingual text-only data, which is normally easy to collect, for pre-training and text-based domain adaptation.

\section*{Acknowledgments}
Keqi Deng is funded by the Cambridge Trust. This work has been performed using resources provided by the Cambridge Tier-2 system operated by the University of Cambridge Research Computing Service (www.hpc.cam.ac.uk) funded by EPSRC Tier-2 capital grant EP/T022159/1.
\bibliography{ref, custom}

\clearpage
\appendix

\begin{table*}[t!] 
\centering
\setlength{\tabcolsep}{2.0mm}
\renewcommand\arraystretch{0.99}
\begin{tabular}{ l|c|c }
\Xhline{2\arrayrulewidth}
 &\multicolumn{2}{c}{Fisher-CallHome Spanish (FSC) (Es-En)} \\
\hline
Domain&\multicolumn{2}{c}{Spontaneous Conversation}\\
Train set &\multicolumn{2}{c}{Fisher-train}\\
\ \ -Duration&\multicolumn{2}{c}{171.6 hours}\\
\ \ -English words&\multicolumn{2}{c}{1441K}\\
\cline{2-3}
Intra-domain test sets&\multicolumn{1}{c|}{Fisher-dev / -dev2 / -test} &CallHome-devtest / -evltest\\
\ \ -Duration &\multicolumn{1}{c|}{4.6 / 4.7 / 4.5 hours}&{3.8 / 1.8 hours}\\
\ \ -English words&\multicolumn{1}{c|}{(avg) 40K / 39K / 39K} &{38K / 19K}\\
\hline
\hline
 &\multicolumn{2}{c}{MuST-C v1.0 En-De} \\
\hline
Domain&\multicolumn{2}{c}{TED Talk}\\
Train set &\multicolumn{2}{c}{train}\\
\ \ -Duration&\multicolumn{2}{c}{400.0 hours}\\
\ \ -German words&\multicolumn{2}{c}{3880K}\\
\cline{2-3}
Test sets&\multicolumn{1}{c|}{tst-COMMON} &tst-HE\\
\ \ -Duration &\multicolumn{1}{c|}{4.1 hours}&{1.2 hours}\\
\ \ -German words&44K &10K\\
\hline
\hline
&\multicolumn{2}{c}{Europarl-ST Es-En}\\
\hline
Domain&\multicolumn{2}{c}{European Parliament}\\
\cline{2-3}
    Cross-domain test sets&dev &test\\
\ \ -Duration &5.4 hours& 5.1 hours\\
\ \ -English words&53K &51K\\
\Xhline{2\arrayrulewidth}
\end{tabular}
\vspace{-0.15cm}
\caption{Statistics of datasets used in this paper}
\label{corpus}
\vspace{-0.2cm}
\end{table*}

\section{Statistics}
\label{sec:appendix:data}
The training and test statistics are shown in Table~\ref{corpus}. The data was pre-processed following standard ESPnet-ST recipes, in which speed perturbation was employed with factors 0.9 and 1.1. 
Following the ESPnet-ST recipes, raw source speech was used as input, and 500 and 4000 BPE \cite{gage1994} were used as the modelling units for FCS and MuST-C En-De, respectively.
Model training was performed on 4 NVIDIA A100 GPUs each with 80GB GPU memory. For the Fisher-CallHome Spanish corpus, each epoch of training took about 26 minutes. For the MuST-C En-De corpus, each training epoch consumed about 80 minutes.

\section{Hyper-parameters and Inference}
\label{hyper}
For the wav2vec2.0 encoder provided by Fairseq \cite{ott2019fairseq}, "xlsr\_53\_56k" was used for FCS data and ``wav2vec\_vox\_new" was used for MuST-C data. SST training was for 35 and 20 epochs for FCS and MuST-C En-De, respectively.
The hyper-parameters of the models we built are as follows, with other hyper-parameters following standard ESPnet-ST recipes.

\paragraph{LS-Transducer-SST}
The LS-Transducer-SST had a 6-layer unidirectional Transformer-encoder-based prediction network (1024 attention dimension, 2048 feed-forward dimension, and 8 heads), resulting in 372.09~M parameters.
The $3$rd sub-layer output of the prediction network was used as the query of the AIF mechanism. $\delta$ in Eq.~\ref{smooth} was set to 0.05.
$\beta$ and $\gamma$ in Eq.~\ref{obj} were respectively set to 0.6 and 0.05. 
The beam size within a chunk was 10 during the chunk-based incremental joint decoding.

\paragraph{Wait-k}
The Wait-k model (404.66~M parameters) had a 6-layer Transformer decoder (1024 attention dimension, 2048 feed-forward dimension, and 8 heads).
The decoding process is similar to the chunk-based incremental decoding of the LS-Transducer-SST. By counting the fixed pre-decision step size and the waiting steps $k$, it is easy to know whether a token is the last one of a chunk,  i.e., whether incremental pruning is needed. 

\paragraph{CIF-IL}
The CIF-IL model (393.63~M parameters) had a 6-layer Transformer decoder (1024 attention dimension, 2048 feed-forward dimension, and 8 heads). 
As explained in Section.\ref{e2e-sst}, the CIF-IL used CIF to estimate when to emit translation tokens and the 6-layer Transformer decoder was built on top of the CIF output.
The decoding process was similar to the incremental decoding of the LS-Transducer-SST because AIF is an improved version of CIF, and they both use frame-level weights to decide when to emit.


\paragraph{CAAT}
\cite{liu-etal-2021-cross} chose block-based streaming Transformer encoder and relied on $d$ to adjust the quality-latency trade-off. However, tuning $d$ is not very suitable for the chunk-based Transformer encoder used in this paper, and preliminary experiments showed that further increasing $d$ did not improve performance, which was also reported in \cite{papi-etal-2022-simultaneous}. Therefore, this paper only sets $d$ to a fixed value of 32.
The attention dimension, feed-forward dimension, and attention heads of the prediction network and the joint network were set to 1024, 2048, and 8. The total parameters are 418.32~M.
During decoding, within each chunk,
beam search was used while chunk-based incremental pruning was performed at the last token, which is easy to decide as the CAAT is still a frame-synchronous model.

\begin{figure}[t!]
    \centering
    \includegraphics[width=75mm]{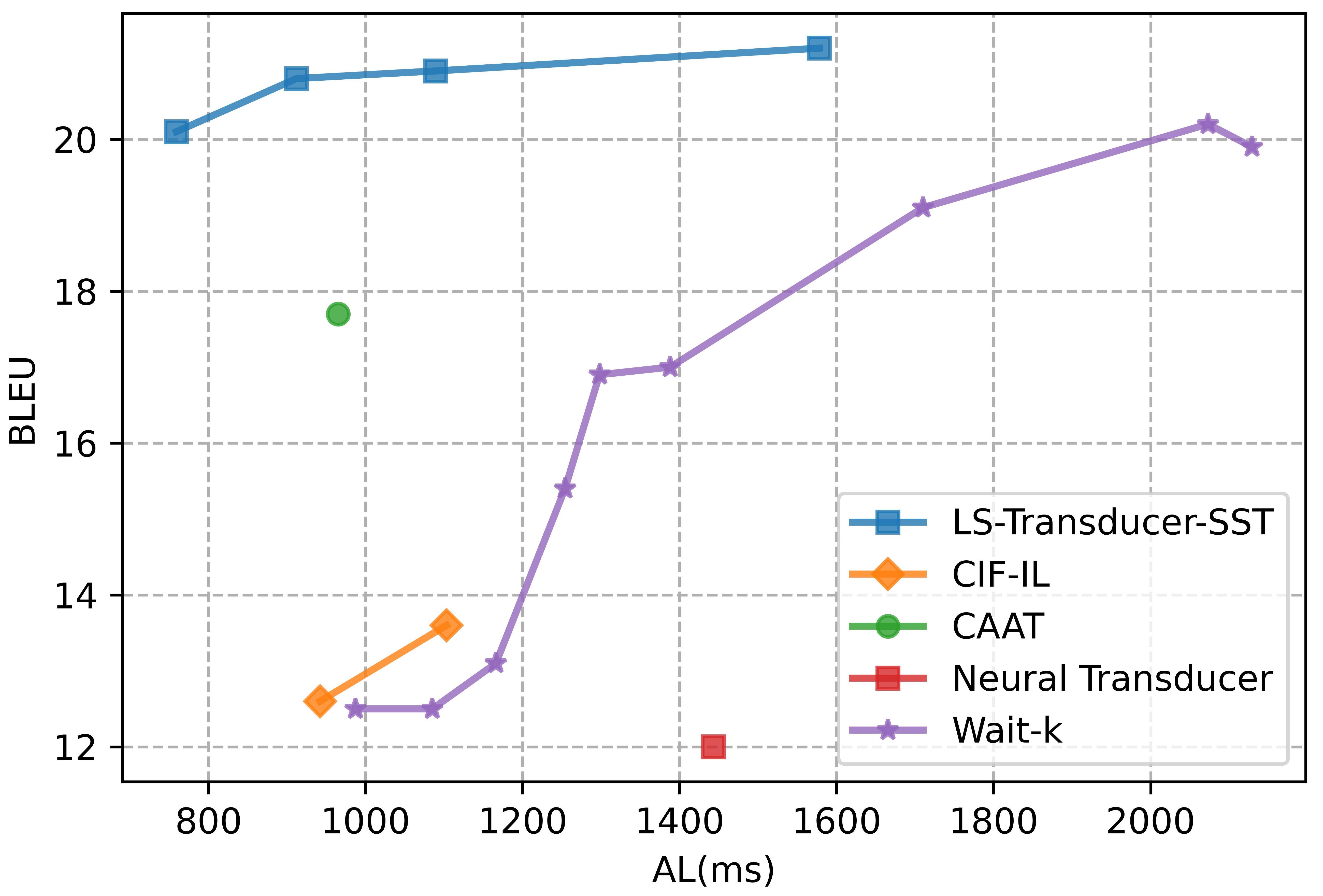}
    \caption{Quality-latency trade-off curves on Fisher-CallHome Spanish CallHome-evltest set, corresponding to Table~\ref{fisher}.}
    \label{tradeoff_fisher}
\end{figure}

\begin{figure}[t!]
    \centering
    \includegraphics[width=75mm]{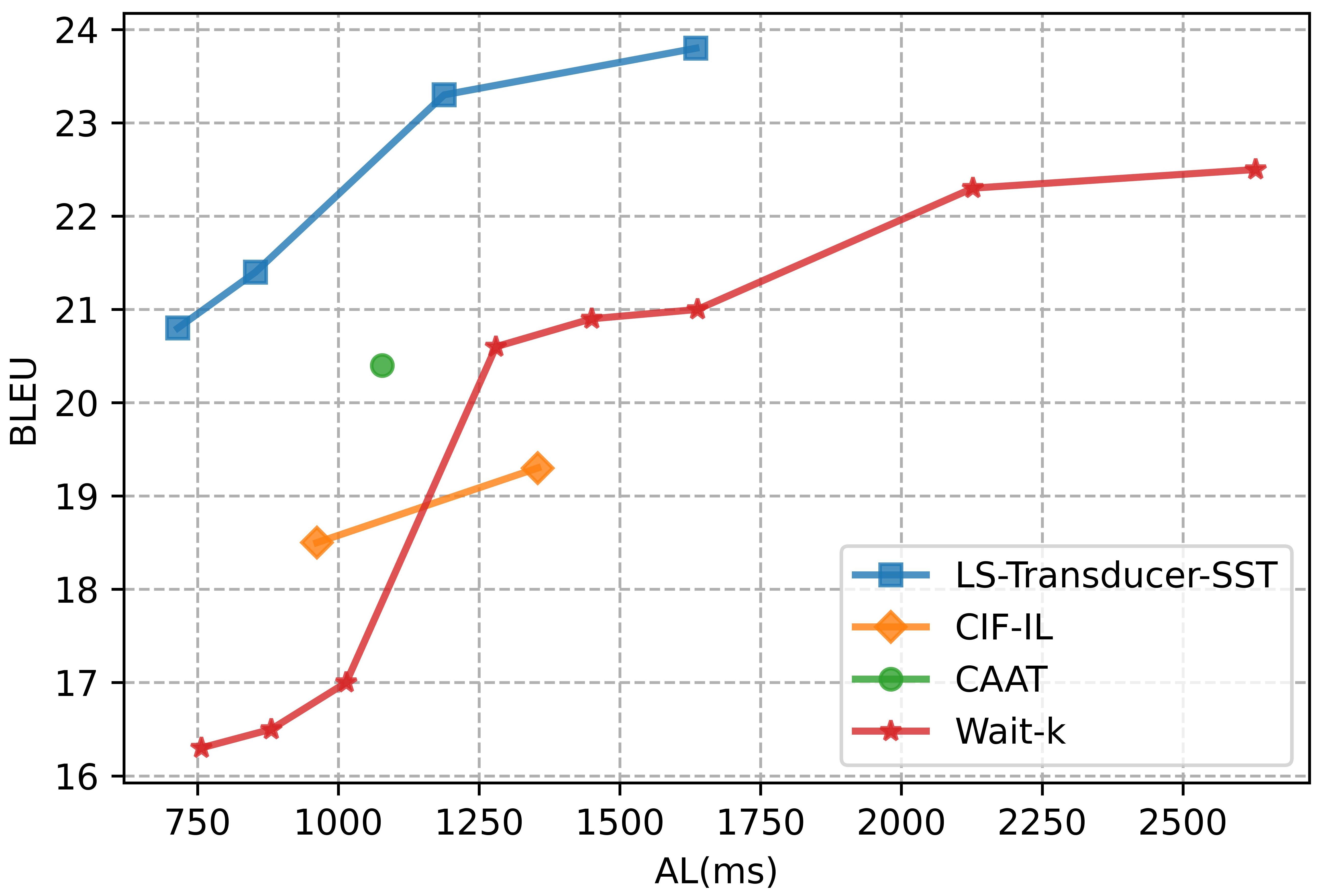}
    \caption{Quality-latency trade-off curves on MuST-C En-De tst-COMMON set, corresponding to Table~\ref{tab:must}.}
    \label{tradeoff_must}
\end{figure}

\paragraph{Standard Neural Transducer}
Decoding was based on beam search with chunk-based incremental pruning. Since the speech is decoded on a per-frame basis, it's easy to know whether a token is the last one of a chunk. The parameters are 369.04~M.

\section{Visualisation: Main Experiments}
\label{vis_main}
The quality-latency trade-off curves in Fig.~\ref{tradeoff_fisher} and Fig.~\ref{tradeoff_must} respectively correspond to the results from Table~\ref{fisher} and Table~\ref{tab:must} of the Main Experiments.

\begin{table}[t] 
  \centering
  \setlength{\tabcolsep}{1.2mm}
  \begin{tabular}{ l | c c| c }
    \Xhline{3\arrayrulewidth}
     \multirow{2}{*}{SST Models}&\multicolumn{2}{c|}{MuST-C En-De}&\multirow{2}{*}{AL(s)}\\
     &{COMMON}&{HE}& \\
    \hline
    LS-Transducer-SST&\textbf{20.8}&\textbf{19.3}&0.715\\
    {\ } w/ tail beam search&18.5&17.4&0.761\\
    {\ } w/ greedy search&17.8&16.8&0.760\\
    \Xhline{3\arrayrulewidth}
  \end{tabular}
  \vspace{-0.1cm}
  \caption{BLEU ($\uparrow$) results for LS-Transducer-SST with different decoding methods. Tail beam search means to use beam search only after reading all the input speech.}
  \label{ablation_chunk}
\end{table}

\begin{table}[t] 
  \centering
  \setlength{\tabcolsep}{0.5mm}
  \begin{tabular}{ l | c c| c }
    \Xhline{3\arrayrulewidth}
     \multirow{2}{*}{SST Models}&\multicolumn{2}{c|}{MuST-C En-De}&\multirow{2}{*}{AL(s)}\\
     &{COMMON}&{HE}& \\
    \hline
    LS-Transducer-SST&\textbf{20.8}&\textbf{19.3}&0.715\\
    {\ } w/ dot-product attention&20.4&18.8&0.710\\
    \Xhline{3\arrayrulewidth}
  \end{tabular}
  \vspace{-0.1cm}
  \caption{BLEU ($\uparrow$) for LS-Transducer-SST with multi-head attention or dot-product attention for the AIF.}
  \label{ablation_att}
  \vspace{-0.2cm}
\end{table}

\section{Further Ablation Studies}
\label{more_ablation}
The results of further ablation studies for the Main Experiments are included in this section to evaluate the effectiveness of the chunk-based incremental joint decoding and the usage of multi-head attention for AIF.

The results in Table~\ref{ablation_chunk} show that using beam search after reading the whole speech (i.e. tail beam search) is better than greedy decoding, and using beam search within each chunk (i.e., chunk-based incremental decoding) performs the best, which it is because it more fully utilises the beam search to widen the search space.

The results in Table~\ref{ablation_att} show that the multi-head attention works better than the simple dot-product attention for the AIF, which is consistent with the success of the Transformer \cite{Vaswani2017}.

\begin{table*}[t]
  \centering
  \setlength{\tabcolsep}{1.2mm}
  \renewcommand\arraystretch{0.91}
  \begin{tabular}{l | c c |c c| c }
    \Xhline{3\arrayrulewidth}
     SST Models&{Fisher-test}&{CallHome-evltest}&LAAL(s)&AL(s)&{Latency}\\
    \hline
    Wait-$5$ with 360~ms &48.9&19.9&2.193&2.129&High\\
    Wait-$1$ with 200~ms &25.2&12.5&1.072&0.987&Low\\
    \hline
    CIF-IL with $\lambda_{lat}=0.0$& 33.1&13.6&1.244&1.103&Medium\\
    CIF-IL with $\lambda_{lat}=0.5$& 30.3&12.6&1.083&0.942&Low\\
    \hline
    CAAT &44.7&17.7&1.103&0.965&Low\\
    \hline
    Standard Neural Transducer &37.9&12.0&1.470&1.443&Medium\\
    \hline
    LS-Transducer-SST with $\epsilon=1$&{47.8}&{20.8}&1.180&{0.912}&Low\\
    LS-Transducer-SST with $\epsilon=5$&{51.4}&{21.2}&1.743&{1.578}&Medium\\
    \Xhline{3\arrayrulewidth}
  \end{tabular}
  \vspace{-0.2cm}
  \caption{Case-sensitive BLEU ($\uparrow$) results on the Fisher-CallHome Spanish. LAAL ($\downarrow$) and AL ($\downarrow$) was tested on the CallHome-evltest. Note that the wav2vec2.0 encoder was used following the main experiment setup.}
  \label{tab:fisher-laal}
  \vspace{-0.1cm}
\end{table*}

\begin{table*}[t!]
  \centering
  \setlength{\tabcolsep}{1.2mm}
  \renewcommand\arraystretch{0.91}
  \begin{tabular}{l | c c |c c| c }
    \Xhline{3\arrayrulewidth}
     SST Models&{COMMON}&{HE}&LAAL(s)&AL(s)&{Latency}\\
    \hline
    Wait-$5$ with 360~ms &22.5&21.9&2.663&2.629&High\\
    Wait-$1$ with 200~ms &16.3&11.2&0.852&0.757&Low\\
    \hline
    CIF-IL with $\lambda_{lat}=0.0$& 19.3&18.2&1.433&1.354&Medium\\
    CIF-IL with $\lambda_{lat}=0.3$& 18.5&17.2&1.081&0.962&Low\\
    \hline
    CAAT &20.4&18.9&1.144&1.078&Medium\\
    \hline
    LS-Transducer-SST with $\epsilon=1$&{21.4}&{20.0}&1.093&{0.853}&Low\\
    LS-Transducer-SST with $\epsilon=5$&{23.8}&{22.3}&1.785&{1.635}&Medium\\
    \Xhline{3\arrayrulewidth}
  \end{tabular}
  \vspace{-0.2cm}
  \caption{Case-sensitive BLEU ($\uparrow$) results on MuST-C En-De. LAAL ($\downarrow$) and AL ($\downarrow$) was tested on the tst-COMMON set. Note that the wav2vec2.0 encoder was used following the main experiment setup.}
  \label{tab:must-laal}
  \vspace{-0.1cm}
\end{table*}

\begin{table*}[t!]
  \centering
  \setlength{\tabcolsep}{1.2mm}
  \renewcommand\arraystretch{0.91}
  \begin{tabular}{l | c c |c c| c }
    \Xhline{3\arrayrulewidth}
     SST Models&{Fisher-test}&{CallHome-evltest}&AL\_CA(s)&AL(s)&{Latency}\\
    \hline
    CAAT &44.7&17.7&1.691&0.965&Low\\
    \hline
    LS-Transducer-SST with $\epsilon=1$&{47.8}&{20.8}&1.435&{0.912}&Low\\
    \Xhline{3\arrayrulewidth}
  \end{tabular}
  \vspace{-0.2cm}
  \caption{Case-sensitive BLEU ($\uparrow$) results on the Fisher-CallHome Spanish. AL\_CA ($\downarrow$) and AL ($\downarrow$) was tested on the CallHome-evltest. Note that the wav2vec2.0 encoder was used following the main experiment setup.}
  \label{tab:fisher-alca}
  \vspace{-0.1cm}
\end{table*}

\begin{table*}[t!]
  \centering
  \setlength{\tabcolsep}{1.2mm}
  \renewcommand\arraystretch{0.91}
  \begin{tabular}{l | c c |c c| c }
    \Xhline{3\arrayrulewidth}
     SST Models&{COMMON}&{HE}&AL\_CA(s)&AL(s)&{Latency}\\
    \hline
    CAAT &20.4&18.9&1.897&1.078&Medium\\
    \hline
    LS-Transducer-SST with $\epsilon=3$&{23.3}&{21.3}&1.882&{1.188}&Medium\\
    \Xhline{3\arrayrulewidth}
  \end{tabular}
  \vspace{-0.2cm}
  \caption{Case-sensitive BLEU ($\uparrow$) results on MuST-C En-De. LAAL ($\downarrow$) and AL ($\downarrow$) was tested on the tst-COMMON set. Note that the wav2vec2.0 encoder was used following the main experiment setup.}
  \label{tab:must-alca}
  \vspace{-0.1cm}
\end{table*}

\section{Expanded Results: Main Experiments}
\label{expand}
In addition to the AL metric, in this section, the results from Table~\ref{fisher} and Table~\ref{tab:must} in the main experiment are additionally measured using the Length-Adaptive Average Lagging (LAAL) \cite{papi-etal-2022-generation} metric.
As shown in Table~\ref{tab:fisher-laal} and Table~\ref{tab:must-laal}, consistent with the findings of \citep{papi-etal-2022-generation}, the Wait-k model tends to have similar AL and LAAL values, indicating a tendency to generate less compared to other models. Overall, the experimental conclusions are consistent for both metrics.

Furthermore, a computation-aware (CA) metric (i.e. AL\_CA) was also considered with A100 GPU. As shown in Table~\ref{tab:fisher-alca} and Table~\ref{tab:must-alca}, the increase from AL to AL\_CA values is smaller for LS-Transducer-SST than for CAAT, indicating that LS-Transducer-SST decodes faster.

\section{Results with Knowledge Distillation}
\label{expand_kd}
Sequence-level knowledge distillation (KD) \cite{kim-rush-2016-sequence} was not used in the main experiment because this is not the default setting of the ESPnet-ST toolkit we used and would double the computational cost. In this section, we select some key data and show the results of using the KD.

\begin{table*}[t]
  \centering
  \setlength{\tabcolsep}{1.2mm}
  \renewcommand\arraystretch{0.95}
  \begin{tabular}{l | c |c  c |c| c }
    \Xhline{3\arrayrulewidth}
     SST Models&Sentence-level KD&{COMMON}&{HE}&AL(s)&{Latency}\\
     \hline
     Wait-$3$ with 280~ms &\XSolidBrush&21.0&18.9&1.638&Medium\\
     Wait-$3$ with 280~ms &\Checkmark&23.8&20.4&1.608&Medium\\
     \hline
     CIF-IL with $\lambda_{lat}=0.0$&\XSolidBrush& 19.3&18.2&1.354&Medium\\
     CIF-IL with $\lambda_{lat}=0.0$&\Checkmark& 21.8&19.6&1.060&Medium\\
     \hline
     CAAT &\XSolidBrush&20.4&18.9&1.078&Medium\\
     CAAT &\Checkmark&23.9&22.5&1.017&Medium\\
    \hline
    LS-Transducer-SST with $\epsilon=1$&\XSolidBrush&{21.4}&{20.0}&{0.853}&Low\\
    LS-Transducer-SST with $\epsilon=1$&\Checkmark&{24.6}&{22.5}&{0.921}&Low\\
    \Xhline{3\arrayrulewidth}
  \end{tabular}
  \vspace{-0.2cm}
  \caption{Case-sensitive BLEU ($\uparrow$) results on MuST-C En-De. AL ($\downarrow$) was tested on the tst-COMMON set. Note that the wav2vec2.0 encoder was used.} 
  \label{tab:must-kd}
  \vspace{-0.1cm}
\end{table*}

As shown in Table~\ref{tab:must-kd}, the sequence-level KD was effective in improving translation quality at the cost of increased training computation.
The experimental conclusions were consistent with and without this KD, i.e., CAAT exceeded
Wait-k and CIF-IL when AL was around 1~s, while the proposed LS-Transducer-SST gave both lower AL latency and higher BLEU scores than other systems.

\section{Training Setting for Section~\ref{related}}
\label{setting}

A chunk-based streaming Conformer \cite{gulati20_interspeech} encoder was built and used only in Section~\ref{related}. 80-dimensional filter bank was computed as the input feature which was computed every 10~ms with a 25~ms window. The speech features were down-sampled by a factor of 4 via two causal convolution layers before being fed into a 12-layer chunk-based streaming Conformer encoder, in which the chunk size was 32. Therefore, the average latency from the chunk-based encoder was 640~ms, which was the same as the main experiment.
The attention dimension, feed-forward dimension and attention heads were set to 512, 2048, and 8.
$\delta$ in Eq.~\ref{smooth} was set to 0.05 as in the main experiment but the resulting $\alpha_t$ was multiplied by 2, considering that here the frame stride of the encoder output was twice that of the main experiment. $\epsilon$ was set to $\{-2,-1,0,1,3\}$ for the latency-controllable AIF.
Sequence-level knowledge distillation \cite{kim-rush-2016-sequence} was built 
using the code provided by \citet{liu-etal-2021-cross}.

\section{Numerical Values for Table~\ref{related_must}}
\label{numeric}
Numerical values for Fig.~\ref{related_must} are shown in Table~\ref{tab:numeric}. In addition, to AL values shown in Fig.~\ref{related_must}, the table also includes the Length-Adaptive Average Lagging (LAAL) \cite{papi-etal-2022-generation} metric.

\begin{table}[t!] 
  \centering
  \setlength{\tabcolsep}{0.6mm}
  \renewcommand\arraystretch{0.97}
  \begin{tabular}{l | c |c c  }
    \Xhline{3\arrayrulewidth}
     SST Models&BLEU&AL(s)&LAAL(s)\\
    \hline
    Wait-k&19.6&1.430&1.530\\
    from \citep{papi-etal-2023-attention}&23.5&2.000&2.100\\
    \hline
    LA&19.5&1.270&1.410\\
    from \citep{papi-etal-2023-attention}&23.1&1.690&1.790\\
    \hline
    CAAT&20.3&0.880&1.020\\
    from \citep{papi-etal-2023-attention}&20.8&1.320&1.400\\
    &20.5&1.740&1.780\\
    \hline
    CAAT&20.5&0.508&---\\
    from \citep{liu-etal-2021-cross}&21.4&0.814&---\\
    &21.8&1.115&---\\
    &22.2&1.443&---\\
    &22.4&1.801&---\\
    \hline
    EDA{\small TT}&16.8&0.880&1.080\\
    from \citep{papi-etal-2023-attention}&19.1&1.040&1.200\\
    &21.6&1.340&1.460\\
    &24.0&1.740&1.830\\
    \hline
    TAED&21.6&1.263&1.411\\
    from \citep{tang-etal-2023-hybrid}&22.6&1.354&1.530\\
    &23.5&1.370&1.544\\
    &23.5&1.903&2.024\\
     \Xhline{3\arrayrulewidth}
    \hline
    LS-Transducer-SST&21.8&0.778&1.037\\
    (fair comparison with &22.2&0.927&1.157\\
    other techniques above)&23.1&1.082&1.288\\
    &23.7&1.256&1.437\\
    &24.2&1.687&1.840\\
    \hline
    LS-Transducer-SST&23.6&0.751&1.025\\
    with wav2vec2.0&24.6&0.922&1.159\\
    (unfair comparison with&25.6&1.294&1.492\\
    other techniques above)&26.1&1.686&1.841\\
     \Xhline{3\arrayrulewidth}
  \end{tabular}
  \vspace{-0.25cm}
  \caption{Case-sensitive BLEU ($\uparrow$) results on MuST-C En-De. AL ($\downarrow$) was tested on the tst-COMMON set. Numeric values corresponding to Fig.~\ref{related}.}
  \label{tab:numeric}
  \vspace{-0.15cm}
\end{table}

\section{Visual Case Study}
\label{app:case}
\subsection{MuST-C En-De}
\label{app:case:ende}
\begin{figure*}[t!]
    \centering
    \includegraphics[width=131mm]{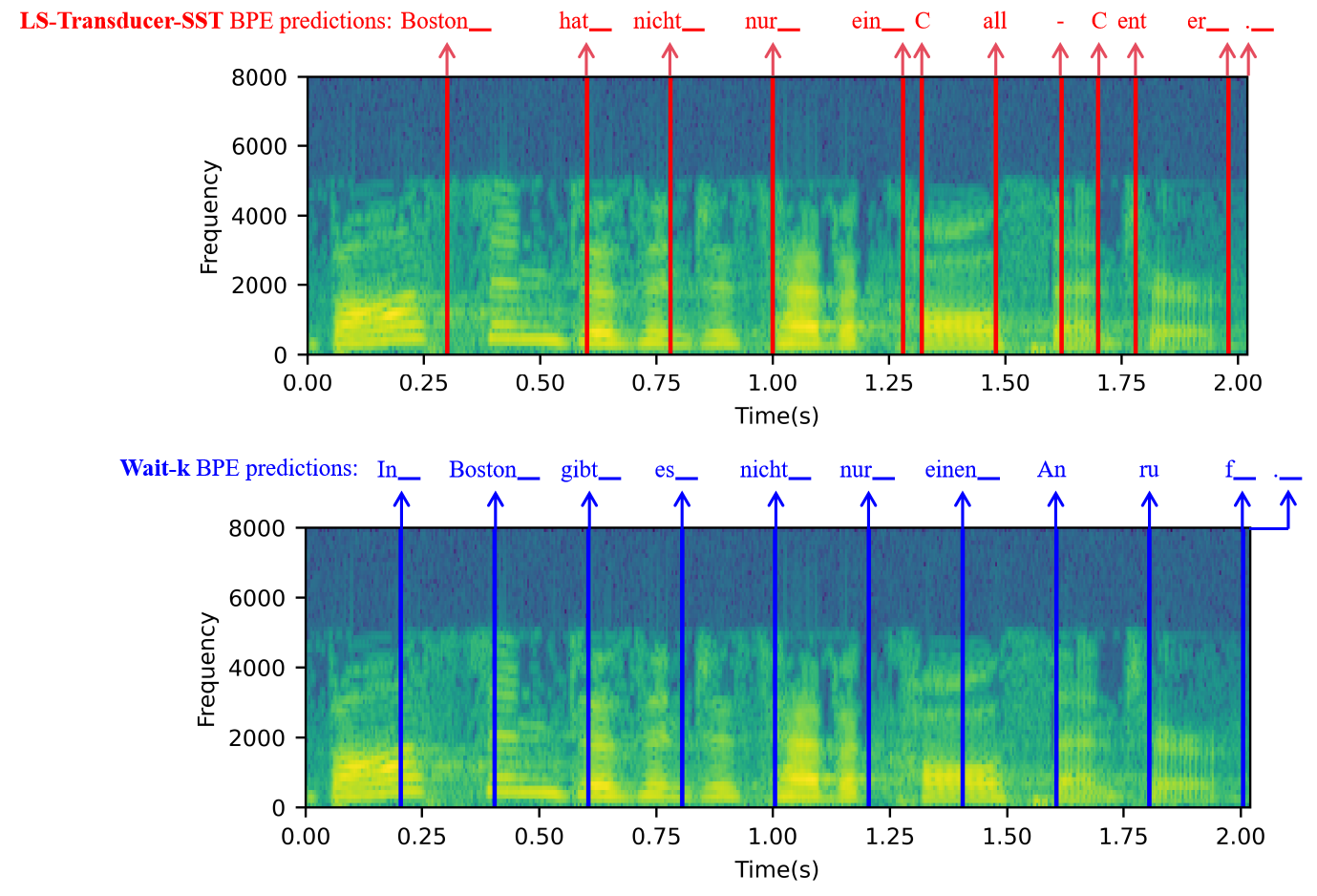}
    \vspace{-0.3cm}
    \caption{An example (377.95s-379.97s segment of ``ted\_01381” from MuST-C En-De tst-COMMON set) of LS-Transducer-SST and Wait-k (Wait-1 with 200~ms). The ground-truth transcript of this utterance is ``\textit{boston doesn’t just have a call center}", while the ground-truth translation is ``\textit{Boston hat nicht nur ein Call-Center .}". The red arrows denote the time steps of the BPE units predicted by the LS-Transducer-SST. The blue arrows represent the time steps for the Wait-k model.}
    \vspace{-0.2cm}
    \label{case4}
\end{figure*}

A visual example from the MuST-C En-De tst-COMMON set is given in Fig.~\ref{case4} to show the re-ordering capability of the LS-Transducer-SST, where the LS-Transducer-SST successfully output ``\textit{hat nicht nur}" that corresponds to the English transcripts ``\textit{doesn't just have}". Note the chunk size of the streaming Transformer encoder is 64, which means a range of 1.28~s.

\subsection{Cross-domain Europarl-ST Es-En}
\label{app:case:esen}
\begin{figure*}[t]
    \centering
    \includegraphics[width=131mm]{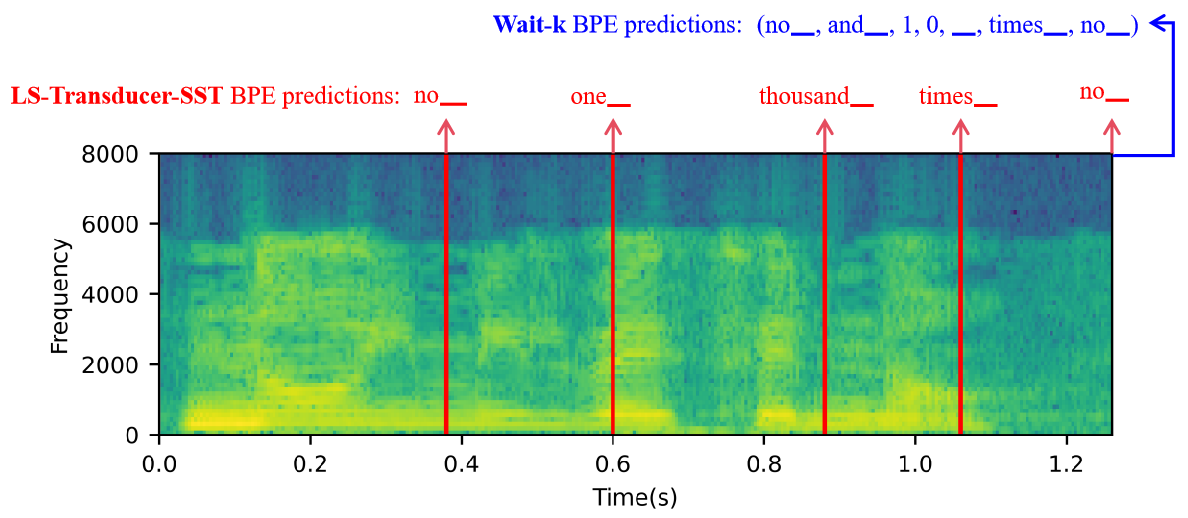}
    \vspace{-0.3cm}
    \caption{An example (65.65s-64.38s segment of ``en.20110704.25.1-169-000” from Europarl-ST Es-En test set) of LS-Transducer-SST and Wait-k (Wait-5 with 360~ms). The ground-truth transcript of this utterance is ``\textit{no y mil veces no}", while the ground-truth translation is ``\textit{no a thousand times no}".}
    \vspace{-0.2cm}
    \label{case}
\end{figure*}

\begin{figure*}[t!]
    \centering
    \includegraphics[width=131mm]{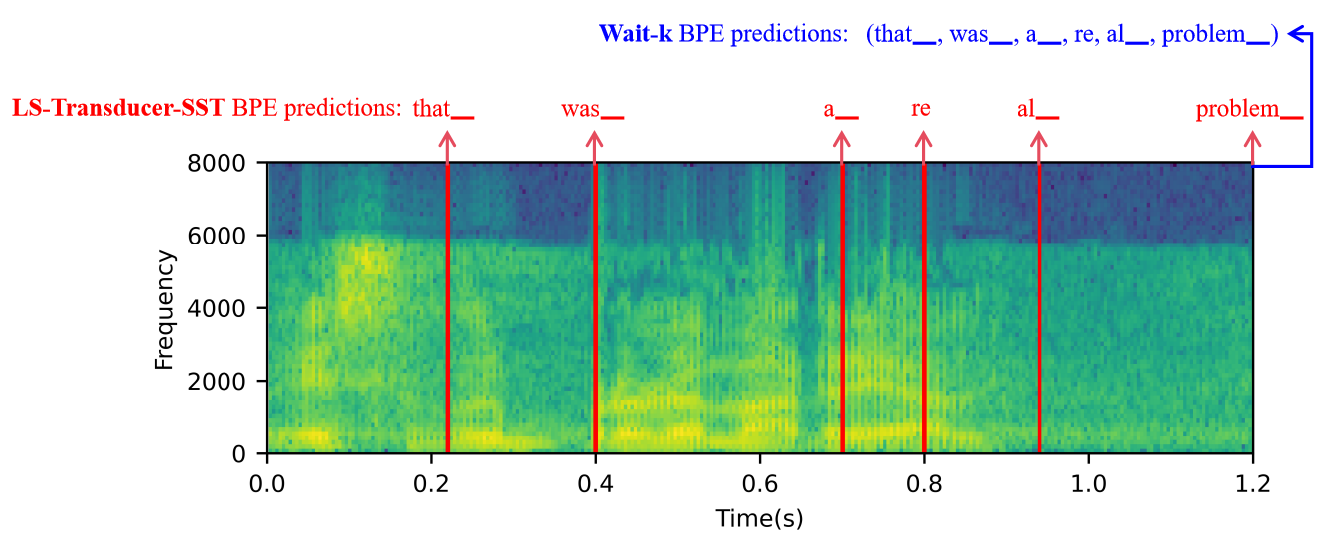}
    \vspace{-0.3cm}
    \caption{An example (85.72s-86.93s segment of ``en.20100120.5.3-063” from Europarl-ST Es-En test set) of LS-Transducer-SST and Wait-k (Wait-5 with 360~ms). The ground-truth transcript of this utterance is ``\textit{es un problema real}", while the ground-truth translation is ``\textit{it is a real problem}".}
    \vspace{-0.2cm}
    \label{case2}
\end{figure*}

A visual case study is given to compare the LS-Transducer-SST and the Wait-k in the cross-domain Europarl-ST Es-En test set. As shown in Fig.~\ref{case}, the Wait-k model has read the whole speech input from the first BPE unit, because $18\times5\times0.02=1.8~s$ (0.02~s is the frame stride), which is greater than the 1.27~s length of the speech. This is a simple example without re-ordering, where the LS-Transducer-SST always outputs each translation token after reading the corresponding source-language speech.
Another example in Fig.~\ref{case2} from the Europarl-ST Es-En test set contains a re-ordering case (``\textit{problema real}" to ``\textit{real problem}"). The LS-Transducer-SST decided to output the English word ``\text{real}" until reading the corresponding source speech.

\end{document}